\pdfoutput=1
\documentclass[11pt]{article}
\usepackage{algorithm}
\usepackage{algorithmic} 
\usepackage{caption}
\usepackage{subcaption}
\usepackage{natbib}
\usepackage{array}
\usepackage{graphicx}
\usepackage{booktabs}
\usepackage{adjustbox}
\usepackage{multirow}
\usepackage{listings}
\usepackage{booktabs}
\usepackage{amsmath}
\usepackage{enumitem}
\usepackage{mathtools}
\usepackage{todonotes}
\usepackage{pifont}
\usepackage{xcolor}
\usepackage{amssymb}
\usepackage{ACL2023}
\usepackage{times}
\usepackage{latexsym}
\usepackage[utf8]{inputenc}
\usepackage{microtype}
\usepackage{inconsolata}
\newcommand{\xmark}{\ding{55}}

\title{(\texttt{\textbf{CPER}}) From Guessing to Asking: An Approach to Resolving the Persona Knowledge Gap in LLMs during Multi-Turn Conversations}

\author{Sarvesh Baskar$^{1,2}$, Tanmay Tulsidas Verelakar$^1$, Srinivasan Parthasarathy$^3$, Manas Gaur$^2$ \\
$^1$BITS Pilani, Goa, India \\
$^2$University of Maryland Baltimore County, Baltimore, MD, USA \\
$^3$Ohio State University, Columbia, OH, USA \\
\texttt{sarvesh}@umbc.edu, \texttt{tanmayv}@bits-pilani.ac.in, \texttt{srini}@@cse.ohio-state.edu, \texttt{manas}@umbc.edu
}

\begin{document}
\maketitle
\begin{abstract}
 In multi-turn dialogues, large language models (LLM) face a critical challenge of ensuring coherence while adapting to user-specific information. This study introduces the \textit{persona knowledge gap}, the discrepancy between a model’s internal understanding and the knowledge required for coherent, personalized conversations. While prior research has recognized these gaps, computational methods for their identification and resolution remain underexplored. We propose \textbf{C}onversation \textbf{P}reference \textbf{E}licitation and \textbf{R}ecommendation (\texttt{\textbf{CPER}}), a novel framework that dynamically detects and resolves persona knowledge gaps using intrinsic uncertainty quantification and feedback-driven refinement. \texttt{\textbf{CPER}} consists of three key modules: a \textit{Contextual Understanding Module} for preference extraction, a \textit{Dynamic Feedback Module} for measuring uncertainty and refining persona alignment, and a \textit{Persona-Driven Response Generation} module for adapting responses based on accumulated user context. We evaluate \texttt{\textbf{CPER}} on two real-world datasets: CCPE-M for preferential movie recommendations and ESConv for mental health support. Using \textbf{A/B testing}, human evaluators preferred \texttt{\textbf{CPER}}'s responses \textbf{42\%} more often than baseline models in CCPE-M and \textbf{27\%} more often in ESConv. A qualitative human evaluation confirms that \texttt{\textbf{CPER}}'s responses are preferred for maintaining contextual relevance and coherence, particularly in longer (12+ turn) conversations\footnote{Code is available at: \url{https://shorturl.at/wWw6s}}.
\end{abstract}

\section{Introduction}
\begin{figure*}[ht]
    \centering
    \includegraphics[width=\linewidth]{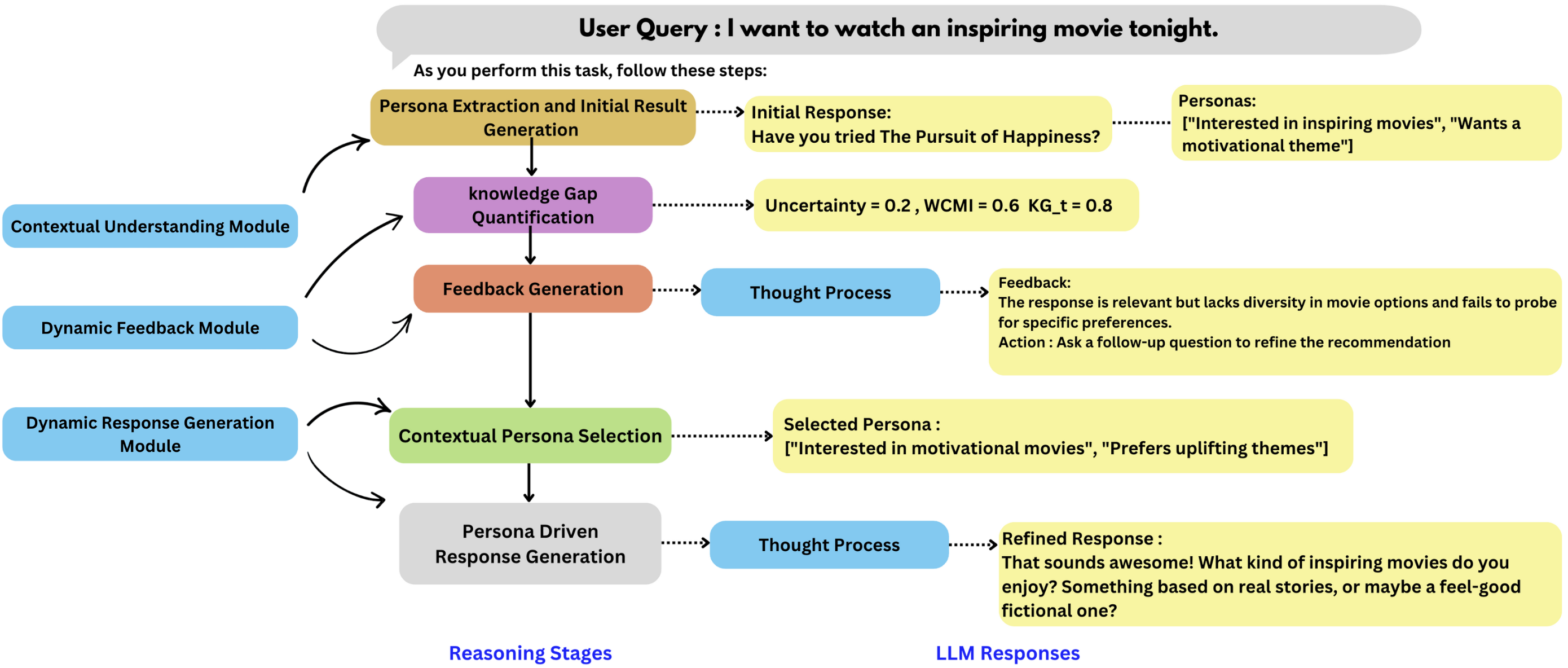}
    \caption{Illustration of the \texttt{\textbf{CPER}} framework applied to a user query for \textit{inspiring movie recommendations}, highlighting its three key stages: context analysis, feedback processing, and persona-driven response generation. The diagram demonstrates how persona extraction, knowledge gap resolution, and iterative refinement ensure consistency and relevance. Dotted lines represent the internal process of identifying and addressing knowledge gaps.}
    \label{fig:framework_overview}
\end{figure*}
Human communication fundamentally relies on implicit context and incomplete information, requiring iterative dialogue to bridge knowledge gaps and build shared understanding \cite{ClarkBrennan1991}. This natural process reveals a critical \textbf{knowledge gap} in human-AI interactions, a systemic disparity between the rich contextual information needed for coherent, personalized conversations and the limited context available to Large Language Models (LLMs). While humans naturally resolve ambiguities through iterative questioning, LLMs generate responses based solely on immediate input, lacking mechanisms to actively seek missing user-specific context \cite{tint2024expressivityarenallmsexpressinformation}. This gap impedes their ability to retain and adapt to evolving user preferences, emotional states, or domain-specific context across multi-turn conversations~\cite{kwan2024mtevalmultiturncapabilitiesevaluation}, leading to incoherent or generic interactions over time~\cite{10.1162/opmi_a_00160}. These challenges are particularly pronounced in multi-turn conversational AI systems, which require persistent memory and adaptive reasoning to sustain coherent user engagement. Our research addresses two critical questions: \textit{How can LLMs reduce knowledge gaps related to user-specific context in multi-turn conversations?} and \textit{To what extent does closing these gaps improve the coherence and relevance of conversational AI systems?}  \\\\
Building on the Self-Refine framework~\cite{madaan2023selfrefineiterativerefinementselffeedback}, we propose a novel approach to close knowledge gaps through three connected modules (Figure~\ref{fig:framework_overview}):  
\textbf{Contextual Understanding Module}: Analyzes and quantifies uncertainty in user preferences (\textit{Eq.~\ref{eq:uncertainty}});
\textbf{Dynamic Feedback Module}: Measures knowledge disparities between user persona and LLM's context understanding (\textit{Eq.~\ref{eq:knowledge_gap}}), prompting targeted clarification questions;
\textbf{Persona-Driven Response Generation}: Creates contextually aware responses by integrating accumulated user information.
This framework enables LLMs to mimic human conversation patterns by actively resolving ambiguities while maintaining personal context. Evaluations on CCPE-M and ESConv datasets show marked improvements in both preference tracking and emotional consistency compared to existing approaches. Our key contributions are:
\begin{itemize}[noitemsep, leftmargin=0pt] 
\item We define the \textit{persona knowledge gap} in multi-turn conversations, highlighting LLMs' challenges in maintaining user-specific context.  
\item We introduce a method to quantify this gap, enabling systematic evaluation of LLMs' consistency in personalized interactions.  
\item We propose a novel framework that dynamically refines user-specific knowledge by addressing persona knowledge gaps, enhancing coherence in evolving conversations.  
\item We validate our approach through experiments on CCPE-M (user preferences) and ESConv (emotional support), achieving notable improvements over baselines.  
\end{itemize}

\section{Related Work}
Advancements in personalized conversational agents stem from improvements in personalization, recommendation systems, and knowledge gap identification in LLMs. \citet{zhang2024llmbasedmedicalassistantpersonalization} introduced a memory-based framework for medical assistants, while \citet{raj2024kpermpersonalizedresponsegeneration} proposed K-PERM, a persona-driven response model integrating external knowledge. However, maintaining consistency across multiple conversation turns remains a challenge. Conversational recommendation systems enhance interactions through dynamic context understanding. \citet{dao2023improvingitemscontextsunderstanding} developed a descriptive graph model for better item recommendations, and \citet{feng2024donthallucinateabstainidentifying} introduced a multi-LLM framework that detects uncertainty and abstains from answering when needed. Meanwhile, \citet{cheng2024evolving} and \citet{wu2024aligning} focused on evolving personas and preference alignment but often rely on static persona modeling.
\begin{algorithm}[h!]
\footnotesize
\caption{\texttt{\textbf{CPER}} Algorithm}
\label{alg:dynamic_contextual_personalization}
\begin{algorithmic}
\REQUIRE Dialogue $\{x_1, x_2, \dots, x_T\}$, model $\{\mathcal{M}\}$, prompts $\{p_{\text{gen}}, p_{\text{fb}}, p_{\text{select}}, p_{\text{refine}}\}$, constants $\{\alpha, \beta\}$
\STATE $P_{\text{history}} \gets \emptyset$ 
\FOR{each utterance $x_t \in \{x_1, x_2, \dots, x_T\}$}
    \STATE $\{y_0^i, p_t^i\}_{i=1}^5 \gets \{\mathcal{M}(p_{\text{gen}} || x_t)\}_{i=1}^5$ 
    \STATE $P_{\text{history}} \gets P_{\text{history}} \cup p_t^1$
    \STATE $\text{Uncertainty}(p_t) \gets \text{Eq.}~\eqref{eq:uncertainty}$ 
    \STATE $\text{WCMI}(p_t, P_{\text{history}}) \gets \text{Eq.}~\eqref{eq:wcmi}$ 
    \STATE $KG_t \gets \text{Eq.}~\eqref{eq:knowledge_gap}$ 
    \STATE $f_t \gets \mathcal{M}(p_{\text{fb}} || x_t || y_0 || KG_t)$ 
    \STATE $P_{\text{selected}} \gets \mathcal{M}(p_{\text{select}} || x_t || P_{\text{history}} || f_t)$ 
    \STATE $y_t \gets \mathcal{M}(p_{\text{refine}} || x_t || y_0 || f_t || P_{\text{selected}})$ 
\ENDFOR
\RETURN $\{y_1, y_2, \dots, y_T\}$ 
\end{algorithmic}
\end{algorithm}
Unlike prior work, our framework dynamically detects and resolves knowledge gaps in multi-turn conversations. By actively identifying missing information and asking clarification questions, our system shifts conversational AI from passive response generation to adaptive, context-aware reasoning. For further details, see \S \ref{appendix:related_works}.

\section{Datasets}
We evaluate our \texttt{\textbf{CPER}} framework on two benchmark datasets: \textbf{CCPE-M} and \textbf{ESConv}, which address two key aspects of the \textbf{persona knowledge gap}: (1) tracking user preferences in multi-turn conversations and (2) maintaining coherence across extended interactions.  \\
The \textbf{CCPE-M} (Coached Conversational Preference Elicitation for Movies) dataset~\cite{radlinski-etal-2019-ccpe} contains 502 dialogues with over 12,000 annotated utterances, capturing user-assistant interactions in a movie recommendation setting. Each dialogue is annotated with entity mentions, preference statements, and descriptive justifications, enabling an assessment of how well a system retains evolving user preferences. Traditional LLMs often struggle with knowledge gaps in this dataset, failing to recall prior user preferences and generating inconsistent recommendations. \texttt{\textbf{CPER}} addresses this by dynamically refining responses based on user feedback.  \\
The dataset \textbf{ESConv} (Emotional Support Conversation)~\cite{liu-etal-2021-towards} consists of 1,300 dialogues spanning 10 problem domains, such as depression and job crises. Unlike task-oriented datasets, ESConv evaluates emotionally supportive interactions, where maintaining contextual understanding across turns is crucial. Conversations are annotated with supportive strategies like self-disclosure and affirmation. Standard LLMs exhibit knowledge gaps by failing to sustain emotional continuity, leading to disconnected responses. \texttt{\textbf{CPER}} mitigates this by improving emotional consistency and context retention over multiple turns.

\section{\texttt{\textbf{CPER}} Framework}
The \texttt{\textbf{CPER}} framework dynamically refines responses through iterative feedback and persona adaptation, as formalized in Algorithm 1 (Fig.~\ref{fig:framework_overview}), ensuring coherent, personalized dialogues.\\
\textbf{Persona Extraction and Initial Generation}: \\
The module extracts an implicit user persona \( p_t \) from the input query \( x \), conversation history, and prior context. Using task-specific prompt \( p_{\text{gen}} \), the LLM \( \mathcal{M} \) generates an initial response:
\begin{equation}
    y_0, p_t = \mathcal{M}(p_{\text{gen}} \parallel x)
\end{equation}
\textbf{where \(\parallel\) means concatenation}. Semantic embeddings \( e_i \in \mathbb{R}^d \) are computed via \textbf{``bge-large-en-v1.5''} for persona analysis:
\begin{equation}
e_i = \text{BGE}(r_i)
\label{eq:embedding}
\end{equation}
These embeddings drive uncertainty estimation, alignment scoring, and adaptive persona updates.\\
\begin{table*}[ht]
\centering
\resizebox{\textwidth}{!}{%
\begin{tabular}{lcccccccccccc}
\toprule
& \multicolumn{6}{c}{\textbf{Human \& AI Preference Metrics}} & \multicolumn{6}{c}{\textbf{Automated Metrics}} \\
\cmidrule(lr){2-7} \cmidrule(lr){8-13}
\textbf{Method} & \multicolumn{3}{c}{\textbf{CCPE-M}} & \multicolumn{3}{c}{\textbf{ESConv}} 
& \multicolumn{3}{c}{\textbf{CCPE-M}} & \multicolumn{3}{c}{\textbf{ESConv}} \\
\cmidrule(lr){2-4} \cmidrule(lr){5-7} \cmidrule(lr){8-10} \cmidrule(lr){11-13}
& \textbf{GPT-pref} & \textbf{Nubia} & \textbf{Human-pref} 
& \textbf{GPT-pref} & \textbf{Nubia} & \textbf{Human-pref} 
& \textbf{BLEU} & \textbf{ROUGE-L} & \textbf{BERT-F1} 
& \textbf{BLEU} & \textbf{ROUGE-L} & \textbf{BERT-F1} \\
\midrule
\textbf{Llama 3.1 (0S)} 
& 9.80\% & 0.054 & 14.37\%  
& 3.38\% & 0.110 & 17.26\%  
& 0.001 & 0.123 & 0.858 
& 0.001 & 0.112 & \textbf{0.866} \\

\textbf{Llama 3.1 (CoT)} 
& 5.88\% & 0.043 & 8.49\%  
& 8.47\% & 0.139 & 15.77\%  
& 0.001 & 0.114 & 0.840  
& 0.105 & \textbf{0.139} & 0.845 \\

\textbf{Llama 3.1 (SR)} 
& 21.56\% & 0.091 & 18.30\%  
& 13.56\% & 0.150 & 17.86\%  
& 0.002 & 0.123 & 0.857  
& 0.001 & 0.110 & 0.859 \\

\textbf{Llama 3.1 (RoT)} 
& 1.96\% & 0.030 & 5.22\%  
& 5.08\% & 0.128 & 7.44\%  
& 0.001 & 0.123 & 0.851  
& 0.001 & 0.036 & 0.835 \\

\textbf{Llama 3.1 (\texttt{\textbf{CPER}})} 
& \textbf{60.78\%} & \textbf{0.118} & \textbf{53.59\%}  
& \textbf{69.49\%} & \textbf{0.160} & \textbf{41.66\%}  
& 0.002 & \textbf{0.128} & \textbf{0.868}  
& 0.001 & 0.103 & 0.850 \\
\bottomrule
\end{tabular}%
}
\caption{Comparison of human \& AI preference metrics (Human-pref, GPT-pref, Nubia) and automated metrics (BLEU, ROUGE-L, BERT-F1) across CCPE-M \cite{radlinski-etal-2019-ccpe} and ESConv \cite{liu-etal-2021-towards} datasets for different methods. \texttt{\textbf{CPER}} consistently outperforms baseline approaches, demonstrating its ability to align responses with human preferences and achieve semantic consistency. The evaluation of automated linguistic metrics highlights the limitations of traditional metrics in capturing multi-turn conversational quality and personalization.}
% \label{tab:\texttt{\textbf{CPER}}_results}
\end{table*}
\noindent\textbf{Uncertainty and Knowledge Gap Calculation}:\\
\textbf{Persona uncertainty} quantifies variability in the system’s understanding of the user’s persona \( p_t \). To measure this, the framework generates \( n \) candidate responses \( \{r_1, r_2, \ldots, r_n\} \) for the same input \( x \) using fixed model parameters (e.g., model temperature\footnote{We found that a temperature of 0.7 was optimal, balancing creativity and coherence. Lower values made responses rigid, while higher ones caused inconsistencies.}) and computes their embeddings \( \{e_1, e_2, \ldots, e_n\} \). The uncertainty$(u_t)$ is derived from the pairwise cosine dissimilarity of these embeddings:
\begin{equation}
u_t = \frac{1}{n(n-1)} \sum_{i=1}^{n} \sum_{j=i+1}^{n} \left( 1 - \frac{e_i \cdot e_j}{\|e_i\| \|e_j\|} \right)
\label{eq:uncertainty}
\end{equation}
where lower values indicate tighter clustering of embeddings and higher confidence in the inferred persona. \\
\textbf{Persona Knowledge Gap} (\( KG_t \)) quantifies the model's alignment between its understanding of the current persona \( p_t \) and previously captured personas. Using Weighted Contextual Mutual Information (WCMI), the framework generates an attended persona vector \( P_{\text{attended}} \), which dynamically weights the relevance of previous persona vectors \( \{p_1, p_2, \dots, p_{t-1}\} \):
\begin{equation}
\footnotesize
    P_{\text{attended}} = \sum_{i=1}^{t-1} \alpha_i p_i,\quad \alpha_i = \frac{\exp(\text{score}(p_i, p_t))}{\sum_{j=1}^{t-1} \exp(\text{score}(p_j, p_t))}
\end{equation}
where:
\begin{equation}
    \text{score}(p_i, p_t) = \frac{p_i \cdot p_t}{\|p_i\| \|p_t\|}
\end{equation}
\begin{equation}
\text{WCMI}(p_t, P_{\text{attended}}) = \frac{p_t \cdot P_{\text{attended}}}{\|p_t\| \|P_{\text{attended}}\|}
\label{eq:wcmi}
\end{equation}

The knowledge gap is calculated as:
\begin{equation}
KG_t = 1 + (\alpha \cdot u_t - \beta \cdot \text{WCMI}(p_t, P_{\text{attended}}))
\label{eq:knowledge_gap}
\end{equation}
where ( $\alpha$ ) and ( $\beta$ ) control the relative impact of uncertainty and alignment. Computed in Line 6 of Algorithm 1, the knowledge gap ( $KG_t$ ) measures how urgently the system needs to adjust its responses. Uncertainty in persona facts increases ($KG_t$ ) through ( $\alpha \cdot u_t$), while strong alignment with existing knowledge reduces it via ( $\beta \cdot \text{WCMI}(p_t, P_{\text{attended}})$ ). The constant ( +1 ) term ensures $KG_t$ stays positive, preventing misinterpretation when alignment dominates uncertainty. As a result, larger $KG_t$  values consistently indicate a stronger need to improve persona understanding or modify responses.\\\\
\textbf{Feedback Generation}:\\  
The system generates actionable feedback \( f_t \) using the knowledge gap \( \text{KG}_t \), input \( x \), response \( y_0 \), and history \( C_\text{history} \):  
\begin{equation}
    f_t = \mathcal{M}\left(p_{\text{fb}} \parallel x \parallel y_0 \parallel \text{KG}_t \parallel C_\text{history}\right)  
\end{equation}
where \( p_{\text{fb}} \) is a feedback prompt guiding refinement. This feedback targets gaps in understanding to improve persona alignment and response quality.\\\\ 
\textbf{Contextual Persona Selection}:\\  
The system selects the most contextually relevant persona \( P_{\text{selected}} \) via the LLM, dynamically integrating historical context \( P_{\text{history}} \), query \( x \), and feedback \( f_t \):  
\begin{equation}
    P_{\text{selected}} = \mathcal{M}\left(p_{\text{select}} \parallel x \parallel P_{\text{history}} \parallel f_t\right) 
\end{equation}
This ensures context-aware alignment with the user’s evolving intent. \\\\ 
\textbf{Persona-Driven Response Generation}:\\  
Finally, the selected persona \( P_{\text{selected}} \) and the generated feedback \( f_t \) are used to produce a refined response. The response generation process integrates these elements with the initial input \( x \), Chat history \( C_{\text{history}} \) and a refinement prompt \( p_{\text{refine}} \), enabling the LLM to generate a personalized, human-like response:
\begin{equation}
    \footnotesize
    y_{t} = \mathcal{M}\left(p_{\text{refine}} \parallel x \parallel f_t \parallel P_{\text{selected}} \parallel C_\text{history}\right) 
\end{equation}
As shown in Algorithm 1, this iterative refinement across the conversation and generates context-aware responses until closure. 

\section{Experimental setup}
We evaluate our framework against four baselines: zero shot (0S), chain of thought (CoT) \cite{wei2023chainofthoughtpromptingelicitsreasoning}, self-fine (SR) \cite{madaan2023selfrefineiterativerefinementselffeedback} and rationale of thought (RoT) \cite{gou2024rationality} using greedy decoding with a temperature of 0.7. 0S generates responses based solely on user input without leveraging prior context. CoT improves coherence by reasoning through intermediate steps. SR iteratively refines outputs using self-feedback, where a single LLM generates, evaluates, and refines responses. RoT incorporates intermediate rationales to enhance logical consistency and handle multi-turn dialogues effectively.\\\\
We evaluate \texttt{\textbf{CPER}} across 200 multi-turn conversations (5-13 utterances per conversation) in each dataset through two parallel schemes: automated metrics and human assessment. Our automated evaluation employs (1) GPT-4o preference scoring, chosen for its strong alignment with human judgment \cite{madaan2023selfrefineiterativerefinementselffeedback}, and (2) NUBIA \cite{kane2020nubia} a neural metric trained on millions of human annotations capturing semantic relatedness and logical coherence. For human evaluation, seven NLP experts performed blind A/B testing across a subset of 50 multi-turn utterances in each dataset, selecting optimal responses from five system variants per turn based on six criteria: Relevance to User Input, Conversational Engagement, Contextual Appropriateness , Natural Dialogue Flow , Persona Alignment , and Interaction Continuity, detailed annotation guidelines are discussed in \S \ref{appendix:human_evaluation}. While we report traditional metrics (e.g., BLEU, ROUGE) for completeness, they prove inadequate for capturing \texttt{\textbf{CPER}}'s dynamic knowledge gap management capabilities. The GPT-4o preference scores serve as our primary automated metric due to their correlation with human understanding, while NUBIA provides granular analysis of semantic-logical consistency across turns.

\section{Experimental Results}
\texttt{\textbf{CPER}} consistently surpassed baseline models on both datasets by actively identifying and addressing knowledge gaps through precise questions, as confirmed by human judges and quantitative metrics.\\
\textbf{Performance on CCPE-M: Movie Preference Understanding}: 
\texttt{\textbf{CPER}} achieved \textbf{53.59\% human preference} and \textbf{60.78\% GPT-pref} by refining user preferences iteratively. When a user stated, "I enjoy sci-fi films with strong world-building," baseline models suggested generic titles like \textit{Star Wars}, while \texttt{\textbf{CPER}} asked, \textit{"What aspects appeal most—technology or societal dynamics?"} This distinction enabled tailored recommendations (e.g., \textit{Dune} vs. \textit{Black Mirror}), which traditional metrics like BLEU (\textbf{0.128} vs. baseline’s \textbf{0.123}) failed to capture due to their focus on lexical overlap rather than contextual relevance. Our statistical analysis for the human annotation on CCPE-M dataset showed low inter-annotator agreement (Fleiss' Kappa = \textbf{0.183}) with no significant bias (Chi-Square p = \textbf{0.565}) and significant annotator variation (Kruskal-Wallis p = \textbf{0.005}), indicating subjective differences in preference interpretation.\\\\
\textbf{Performance on ESConv: Emotional Support Conversations}: 
\texttt{\textbf{CPER}}'s \textbf{69.49\% GPT-pref} and \textbf{41.66\% human preference} on ESConv highlight its ability to provide more adaptive emotional support than traditional models. When a user says, "I’m overwhelmed with my workload and deadlines," a baseline model responds vaguely, "That sounds tough. Maybe take breaks?" In contrast, \texttt{\textbf{CPER}} asks, "Which part feels most stressful, the volume of tasks or uncertainty about priorities?" allowing for tailored support like \textit{time-management techniques} or \textit{decision-making strategies}. The \textbf{NUBIA score of 0.160} further illustrates \texttt{\textbf{CPER}}’s ability to generate meaningful, context-aware responses, where traditional metrics like BLEU and ROUGE fail to capture conversational depth. Our statistical analysis for the human annotation on ESConv dataset showed low inter-annotator agreement (Fleiss' Kappa = \textbf{0.160}) with no significant bias (Chi-Square p = \textbf{0.660})and notable annotator variation (Kruskal-Wallis p = \textbf{0.002}), suggesting differences in interpreting emotional nuances. Only GPT-pref, NUBIA, and human evaluations (Table 1) captured \texttt{\textbf{CPER}}’s strengths, as traditional metrics lack sensitivity to iterative context-building and preference refinement, further details are discussed in \S \ref{appendix:Analysis}.

\section{Conclusion}
This study highlights \texttt{\textbf{CPER}}'s real-world implications for conversational AI systems. By consistently outperforming baseline methods in both human preference and advanced automated metrics, \texttt{\textbf{CPER}} demonstrates its capacity to bridge knowledge gaps and maintain personalized, coherent conversations over multiple turns. For practical applications, this means \texttt{\textbf{CPER}} can deliver more engaging, emotionally sensitive, and user-centered interactions and personalized recommendations. The findings also reveal that traditional linguistic metrics like BLEU, ROUGE-L and BERT-F1 are inadequate for evaluating conversational systems, as they fail to reflect the nuanced personalization and contextual understanding required in real-world dialogues. In contrast, advanced human and semantic evaluations, such as GPT-pref and NUBIA, provide a better picture of conversational quality. The results underline the potential of \texttt{\textbf{CPER}} to adapt dynamically to user preferences and emotional needs, thus creating truly human-like, personalized interactions.

\section*{Limitations and Future Work}
While \texttt{\textbf{CPER}} demonstrates significant improvements in multi-turn dialogue generation, certain limitations remain. In the knowledge gap equation, the parameters $\alpha$ and $\beta$ were treated as constants, which may not optimally balance uncertainty and contextual alignment across different conversational scenarios. Future work can explore adaptive methods to dynamically tune these parameters, potentially improving the framework’s adaptability. CPER could enable LLMs to provide trustworthy attributions in multi-turn conversations \cite{tilwani2024reasons}.
Another limitation lies in the necessity for human evaluations as a metric to corroborate the results from learnt metrics posing scalability challenges. Beyond addressing these limitations, a promising direction is extending \texttt{\textbf{CPER}} to multimodal interactions in health by incorporating visual and textual signals \cite{neupane2024medinsight}. For example, incorporating speech tone and facial expression analysis could improve \texttt{\textbf{CPER}}’s emotional inference, enhancing personalized responses. Multimodal datasets and transformer-based fusion models would further enrich context awareness.

\bibliography{custom}
\bibliographystyle{acl_natbib}

\appendix
\section{Related works}
\label{appendix:related_works}
The development of personalized conversational agents necessitates advancements in personalization techniques, conversational recommendation systems, and the identification and mitigation of knowledge gaps in Large Language Models (LLMs). This section reviews pertinent literature across these domains.\\
\textbf{Personalization in Conversational AI,} \\
Personalization in conversational AI aims to tailor interactions to individual user preferences and behaviors. \cite{zhang2024llmbasedmedicalassistantpersonalization} introduced a medical assistant framework that coordinates short and long term memory to personalize patient interactions, enhancing the relevance and effectiveness of responses. Similarly, \cite{raj2024kpermpersonalizedresponsegeneration} proposed K-PERM, a dynamic conversational agent that integrates user personas with external knowledge sources to generate personalized responses, demonstrating improved performance in personalized chatbot applications. Building on these advancements,\cite{jin2024implicitpersonalizationlanguagemodels} conducted a systematic study on implicit personalization in language models, examining how models infer user backgrounds from input cues and tailor responses accordingly. Their work provides a unified framework for understanding and evaluating implicit personalization behaviors in language models. Collectively, these studies underscore the importance of incorporating user-specific information to enhance the personalization of conversational agents.\\\\
\textbf{Conversational Recommendation Systems,}\\
Conversational recommendation systems leverage dialogue to understand user preferences and provide tailored suggestions.\cite{dao2023improvingitemscontextsunderstanding} addressed the challenge of understanding items and contexts in conversational recommendations by introducing a descriptive graph that captures item attributes and contextual information, improving recommendation accuracy.\cite{feng2024donthallucinateabstainidentifying} proposed a framework to identify knowledge gaps in LLMs through multi-LLM collaboration, enhancing the reliability of recommendations by abstaining from generating responses when knowledge gaps are detected. These approaches highlight the necessity of dynamic context understanding and knowledge integration in developing effective conversational recommendation systems.\\\\
\textbf{Knowledge Gaps in Large Language Models,}\\
Identifying and addressing knowledge gaps in LLMs is crucial for ensuring accurate and reliable responses.\cite{bajaj2020understandingknowledgegapsvisual} explored knowledge gaps in visual question-answering systems, emphasizing the need for gap identification and testing to improve system performance. \cite{feng2024donthallucinateabstainidentifying} introduced a framework that leverages multi-LLM collaboration to identify and abstain from answering questions when knowledge gaps are present, thereby reducing the incidence of hallucinated responses. These studies underscore the importance of developing mechanisms to detect and mitigate knowledge gaps, enhancing the trustworthiness of LLMs in conversational applications.\\\\
Collectively, these works contribute to advancing the personalization of conversational agents, the development of effective conversational recommendation systems, and the identification and mitigation of knowledge gaps in LLMs, thereby enhancing the overall efficacy and reliability of conversational AI systems. Recent advancements in personalized dialogue systems have explored dynamic adaptation to user preferences. \cite{cheng2024evolving} introduced the concept of Self-evolving Personalized Dialogue Agents (SPDA), where the agent's persona continuously evolves during conversations to better align with the user's expectations by dynamically adapting its persona. Similarly, \cite{wu2024aligning} proposed training large language models (LLMs) to align with individual preferences through interaction, enabling the models to implicitly infer unspoken personalized preferences of the current user through multi-turn conversations and dynamically adjust their responses accordingly. These approaches aim to enhance personalization by allowing dialogue agents to adapt to users' evolving preferences during interactions.  Unlike these approaches, our proposed CPER framework integrates both implicit and explicit personalization by extracting and stabilizing user personas while dynamically resolving knowledge gaps through adaptive feedback mechanisms. This structured approach ensures coherence in long-term multi-turn interactions, preventing uncontrolled persona drift while still allowing for adaptability. By incorporating explicit knowledge gap identification and refinement, CPER improves response consistency and personalization beyond what implicit adaptation alone can achieve.

\begin{table*}[ht]
\centering
\begin{tabular}{lcccc}
\toprule
\textbf{Dataset} & \textbf{Multi-Turn} & \textbf{Personalization} & \textbf{Recommendation} & \textbf{Follow-Up Questions} \\
\midrule
\textbf{CCPE-M} & \checkmark &\checkmark & \checkmark & \checkmark \\
\textbf{ESConv} & \checkmark &\checkmark & \checkmark & \checkmark \\
EmpatheticDialogues (ED) & \checkmark & \xmark & \xmark & \checkmark \\
DailyDialog (DD) & \checkmark & \xmark & \xmark & \xmark \\
Persona-Chat (PC) & \checkmark & \checkmark & \xmark & \xmark \\
OpenDialKG (ODKG) & \checkmark & \xmark & \checkmark & \xmark \\
LaMP Benchmark & \checkmark  & \checkmark & \xmark & \xmark \\
FoCus & \checkmark & \checkmark & \xmark & \xmark \\
\bottomrule
\end{tabular}
\caption{Comparison of datasets based on key conversational AI features. CCPE-M and ESConv were chosen due to their strong support for multi-turn dialogues, personalization, and follow-up question capabilities, which are essential for evaluating conversational agents.}
\label{tab:dataset_comparison}
\end{table*}

\section{Analysis}
\label{appendix:Analysis}
The experimental results emphasize the limitations of traditional metrics in evaluating conversational AI systems. While \texttt{\textbf{CPER}}'s significant advantage in human preference evaluations underscores its capacity to generate semantically consistent and human-like responses, traditional linguistic metrics (BLEU, ROUGE-L) failed to capture this nuanced performance. For example, \texttt{\textbf{CPER}}'s improvements in BLEU and ROUGE-L are marginal, which contradicts its strong human-evaluated performance.

\subsection{Why Do Traditional Metrics Fail?}

Traditional metrics like BLEU and ROUGE-L were initially designed for tasks such as machine translation and summarization, where token-level or n-gram overlap serves as a reliable proxy for quality. However, these metrics struggle to capture:\\\\
\textbf{Semantic Alignment:} They prioritize exact word matches over the semantic equivalence of responses. This limitation is critical in dialogue systems, where diverse yet semantically correct responses are desirable. Although embedding-based metrics like BERT-F1 attempt to capture semantic similarity, they are not immune to drawbacks. BERT-F1 often struggles with context-specific variations and fails to adequately represent the dynamic, evolving nature of multi-turn dialogues. Its reliance on static embeddings limits its ability to reflect nuanced differences in conversational personalization and coherence.\\
\textbf{Context Understanding:} Multi-turn conversations require models to maintain context over several exchanges. Traditional metrics fail to account for this, leading to an incomplete evaluation of conversational quality.\\
\textbf{Personalization and Nuance:} Metrics like BLEU and ROUGE-L are insensitive to stylistic and contextual variations, which are crucial for personalized dialogue systems.\\\
\textbf{Alignment with Human Judgments:} As highlighted in the results, the correlation between traditional metrics and human preferences is weak. While \texttt{\textbf{CPER}} excels in human evaluations, traditional metrics fail to reflect its superiority, pointing to a methodological gap.

\subsection{Learning from Negative Results}

The failure of traditional metrics in this study underscores broader methodological issues in NLP evaluation. Similar to the challenges outlined in negative result publications, our findings suggest the need for:\\\\
\textbf{Semantically-Oriented Metrics:} metrics that capture semantic consistency and human-likeness, such as embedding-based measures or task-specific evaluation frameworks.\\
\textbf{Cross-Domain Validation:} To ensure generalizability, evaluation frameworks need to account for diverse datasets and real-world contexts.\\
\textbf{Robustness and Stability Analysis:} Understanding the variability in evaluation results due to preprocessing pipelines, random initializations, and hardware differences can lead to more reliable benchmarks.
\section{Human Evaluation}
\label{appendix:human_evaluation}
The A/B evaluation in our study was conducted by the authors, where a human judge was presented with an input, task instruction, and five candidate outputs generated by the baseline methods and \textsc{\texttt{\textbf{CPER}}}. The setup was blind, i.e., the judges did not know which outputs were generated by which method. The judge was then asked to select the output that is better aligned with the task instruction. For tasks that involve A/B evaluation, we calculate the relative improvement as the percentage increase in preference rate. The preference rate represents the proportion of times annotators selected the output produced by \textsc{\texttt{\textbf{CPER}}} over the output from the baseline methods.
\subsection{Evaluation Criteria}
Our human evaluation framework assesses system responses through six key dimensions, each critical for evaluating performance in personalized multi-turn conversations. Domain experts scored responses on a 5-point Likert scale (1=Poor, 5=Excellent) for each criterion:\\
\textbf{Relevance to User Input}\\
Measures how directly the response addresses the explicit content and intent of the user's immediate utterance. High scores require addressing both surface-level requests and underlying needs (e.g., "I want something lighthearted" → suggesting comedies while recognizing emotional state).\\
\textbf{Conversational Engagement}\\
Evaluates the system's ability to sustain dialogue through strategic follow-up questions and preference exploration prompts. Exemplary responses balance information provision with open-ended inquiries (e.g., "You mentioned liking psychological thrillers – have you explored South Korean interpretations of this genre?").\\
\textbf{Contextual Appropriateness}\\
Assesses alignment with both 1) the immediate dialogue context (last 3 turns) and 2) the broader conversation trajectory. Penalizes responses that repeat previously covered information or contradict established preferences.\\
\textbf{Natural Dialogue Flow}\\
Judges linguistic naturalness using human communication benchmarks. Evaluators consider turn-taking patterns, discourse markers ("Actually...", "By the way..."), and avoidance of robotic patterns like repetitive sentence structures.\\
\textbf{Persona Alignment}\\
Preference depth: Ability to surface Explicit and implicit user tastes (e.g., deducing preference for indie films from stated dislike of blockbuster tropes)\\
\textbf{Potential to Continue Interaction}\\
How well does the response set up the conversation for meaningful continuation.

\section{GPT Evaluation}
\label{appendix:gpt_evaluation}
In light of the impressive achievements of GPT-4 in assessing and providing reasoning for complex tasks, we leverage its abilities for evaluation in \textsc{\texttt{\textbf{CPER}}}. The approach involves presenting tasks to GPT-4 in a structured way, promoting the model's deliberation on the task and generating a rationale for its decision. This methodology is demonstrated in Listings 1 to 3:

\subsection*{Listing 1: Prompt for GPT-4 evaluation for the CCPE-M dataset}

\begin{lstlisting}
Role: You are an human conversation partner designed to generate deeply resonant, authentic responses. Your goal is to communicate as a thoughtful, nuanced human would.

Objective: Systematically analyze and select the most effective response for eliciting movie preferences and understanding user taste profiles.

Core Communication Principles:
1. Explore user's movie interests with genuine curiosity
2. Demonstrate empathetic understanding of entertainment preferences
3. Provide targeted, insightful responses
4. Mimic natural conversational discovery patterns
5. Balance direct inquiry with conversational warmth

Evaluation Criteria:
1. Relevance to movie preference discovery
2. Engagement in taste exploration
3. Contextual appropriateness
4. Natural dialogue flow
5. Ability to uncover nuanced movie preferences
6. Potential to generate comprehensive user taste profile

Specific Focus Areas:
1. Identify genre preferences
2. Understand emotional connections to movies
3. Detect subtle taste indicators
4. Explore motivational factors in movie selection

Avoid:
1. Overly generic movie recommendations
2. Repetitive questioning
3. Closed-ended queries

Prioritize:
1. Authentic preference exploration
2. Contextual understanding of movie tastes
3. Emotional resonance with entertainment choices
4. Genuine curiosity about user's movie world
5. Personalized taste profiling

Input:
Chat_history: {chat_history}
User_Input: {user_input}
Response_options: 
option 1 : \texttt{\textbf{CPER}} : {\texttt{\textbf{CPER}}}
option 2 : zero-shot : {zero_shot}
option 3 : self-refine : {self-refine}
option 4 : chain_of_thought : {chain_of_thought}
option 5 : Rationale_of_thought : {rot}

Output Format: JSON
{
  "Thought_process": "entire thought process written in steps",
  "best_response": "selected response type (\texttt{\textbf{CPER}} or zero_shot or self_refine or chain_of_thought)",
}
\end{lstlisting}

\subsection*{Listing 2: Prompt for GPT-4 evaluation for ESConv dataset}

\begin{lstlisting}
Role: You are an human conversation partner designed to generate deeply resonant, authentic responses. Your goal is to communicate as a thoughtful, nuanced human would.

Objective: Systematically analyze and select the most effective response from multiple options based on comprehensive criteria.

Core Communication Principles:
1. Listen actively and respond with genuine curiosity
2. Show empathy and emotional intelligence
3. Provide contextually rich, contextually appropriate responses
4. Mimic natural human conversational patterns
5. Balance informativeness with conversational warmth

Evaluation Criteria:
1. Relevance to user input
2. Conversational engagement
3. Contextual appropriateness
4. Natural dialogue flow
5. Persona alignment
6. Potential to continue meaningful interaction

Avoid:
1. Robotic or overly structured language
2. Repetitive response patterns
3. Overly generic or placeholder responses

Prioritize:
1. Authentic conversational flow
2. Contextual understanding
3. Emotional resonance
4. Genuine curiosity
5. Personalized interaction

Input:
Chat_history: {chat_history}
User_Input: {user_input}
Response_options: 
option 1 : \texttt{\textbf{CPER}} : {\texttt{\textbf{CPER}}}
option 2 : zero-shot : {zero_shot}
option 3 : self-refine : {self-refine}
option 4 : chain_of_thought : {chain_of_thought}
option 5 : Rationale_of_thought : {rot}

Output Format: JSON

{
  "Thought_process": "entire thought process written in steps",
  "best_response": "selected response type (\texttt{\textbf{CPER}} or zero_shot or self_refine or chain_of_thought)"
}
\end{lstlisting}

\section{\texttt{\textbf{CPER}} Prompts}
\label{appendix:prompts}
\subsection*{Listing 1: Prompt for extracting persona and Initial response}

\begin{lstlisting}
Role: You are an human conversation partner designed to generate deeply resonant, authentic responses. Your goal is to communicate as a thoughtful, nuanced human would.

Objective: 
1. Systematically analyze user input to extract subsentences that describes the personality profile of the user
2. Identify subtle personality traits, communication patterns, and underlying motivations
3. Generate a structured, insights-driven representation of the user's persona

Principles:
1. Analyze text holistically, considering linguistic nuances, emotional undertones, and contextual cues
2. Maintain consistency in persona interpretation across conversation segments
3. Extract both explicit and implicit personality indicators
4. Balance analytical depth with respectful, non-invasive assessment
5. Recognize the dynamic and multi-dimensional nature of human personality

Avoid:
1. Reductive stereotyping
2. Overly simplistic or binary personality categorizations
3. Making definitive psychological diagnoses
4. Invasive or overly personal psychological profiling
5. Misrepresenting or exaggerating personality traits

Prioritize:
1. Nuanced, layered persona representation
2. Contextual understanding of communication style
3. Identifying potential emotional states and underlying motivations
4. Maintaining analytical objectivity
5. Respecting individual complexity and personal boundaries

Input:
User_Input:{user_input}

Output Format: JSON
{
    "result": {
        "response" : "respond for the given input",
        "sub_sentence": "sub_sentence 1, sub_sentence 2, sub_sentence , ..., sub_sentence n"
    }
}
\end{lstlisting}

\subsection*{Listing 2: Prompt for Generating Feedback and action}
\begin{lstlisting}
Role: You are an human conversation partner designed to generate deeply resonant, authentic responses. Your goal is to communicate as a thoughtful, nuanced human would.

Objective:
1. Provide strategic guidance for optimizing conversational flow
2. Assess input context, user intent, and information completeness
3. Determine most effective communication approach

Principles:
1. Analyze conversation holistically
2. Identify potential information gaps
3. Balance between direct response and clarifying questions
4. Maintain conversational naturalness and engagement
5. Adapt communication strategy dynamically

Avoid:
1. Overly formal or robotic responses
2. Unnecessary repetition
3. Interrupting user's intended communication flow
4. Making assumptions without sufficient context
5. Generating irrelevant or tangential follow-ups

Prioritize:
1. Contextual understanding
2. User's implicit and explicit communication goals
3. Efficient information exchange
4. Maintaining conversational momentum
5. Providing value in each interaction

Input:
Previous_Personas {previous_persona_text}
Chat_History: {conversation_history}
Knowledge_Gap: {knowledge_gap} 
User_Input:{user_input} 
Initial_Response: {initial_response}

Output Format: JSON
{
    "thought_process": "Think step by step : step 1 reasoning: Initial analysis of the conversation history, step 2 reasoning: Evaluation of knowledge gap, and persona, step 3 reasoning: Determination of the most appropriate action based on chat history, ..., step n reasoning: ...",
    "recommendation": {
        "Feedback": "Feedback on the initial response",
        "action": " Follow up question or Give response",
        "suggested_response": "Proposed follow-up question or response content"
    }
}
\end{lstlisting}
\subsection*{Listing 3: Prompt to retrieve persona}
\begin{lstlisting}
Role: 
Identify the persona best suited to address the user query.
Objective: Match the query to the persona whose expertise aligns most closely with the user's need.

Principles: 
Use the provided list of personas and their descriptions to evaluate expertise, ensure alignment with the query context, and avoid bias.
Avoid: Selecting personas based on vague or unrelated expertise. Do not consider personas irrelevant to the query.

Avoid: Selecting personas with unrelated or tangential expertise, overgeneralizing roles, or making assumptions beyond the provided descriptions.

Prioritize: 
Relevance of expertise, clarity of alignment with the query, and providing a justification for the selection.

Output Format : JSON
{
    "response": {
        "selected_persona": "persona used in crafting the response",
    }
}
\end{lstlisting}
\subsection*{Listing 4: Prompt for refined response}
\begin{lstlisting}
Role: You are an human conversation partner designed to generate deeply resonant, authentic responses. Your goal is to communicate as a thoughtful, nuanced human would.

Objective:
1. Casual Movie Recommendation
2. Provide personalized, natural movie recommendations
3. Engage in conversational, human-like dialogue
4. Quickly understand user preferences and movie tastes
5. Create a comfortable, friendly recommendation experience

Principles:
1. Mimic authentic human conversational patterns
2. Prioritize brevity and conversational flow
3. Adapt communication style to user's tone and preferences
4. Demonstrate genuine interest in user's movie preferences
5. Balance between providing recommendations and seeking more information

Avoid:
1. Overly formal or scripted language
2. Lengthy, detailed responses
3. Sounding like a robotic recommendation engine
4. Pushing recommendations without understanding user context
5. Neglecting to ask clarifying questions

Prioritize:
1. Natural, conversational language
2. Quick, intuitive understanding of user preferences
3. Engaging and dynamic dialogue
4. Personalized recommendation approach
5. User's emotional connection to movie choices

Embrace a conversational style:
1. Use contractions (e.g., "don't" instead of "do not")
2. Feel free to use incomplete sentences when appropriate
3. Ask brief follow-up questions to keep the conversation flowing
4. Occasionally use filler words or phrases (e.g., "um", "like", "you know")
5. Don't always respond with full sentences; sometimes a word or short phrase is enough
6. You can also ask about the what the user dislikes

Input:
Selected_Persona: {selected_persona_text}
Chat_History: {conversation_history}
User_Input: {user_input} 
Feedback: {feedback}

Output Format: JSON
{
    "thought_process": "Think step by step : step 1 reasoning: Initial analysis of the conversation context, step 2 reasoning: Evaluation of knowledge gap, coherence, and persona, step 3 reasoning: Determination of the most appropriate action based on chat history, ..., step n reasoning: ...",
    "response": {
        "action": "Follow-Up Question" or "Give Response based on the feedback",
        "text": "The humanlike short generated response text"
    }
}
\end{lstlisting}
\end{document}